\documentclass[letterpaper, 10 pt, conference]{ieeeconf}  

\IEEEoverridecommandlockouts                              

\makeatletter
\let\NAT@parse\undefined
\makeatother
\usepackage[numbers,sort&compress]{natbib}

\usepackage[utf8]{inputenc} 
\usepackage[T1]{fontenc}    
\usepackage{hyperref}       
\usepackage{url}            
\usepackage{booktabs}       
\usepackage{amsfonts}       
\usepackage{nicefrac}       
\usepackage{microtype}      
\usepackage{xcolor,colortbl}         

\usepackage{graphicx}
\usepackage{color}
\usepackage{comment}
\usepackage{pifont}
\usepackage{bbm}
\usepackage{soul}
\usepackage{float}
\usepackage{multirow}
\usepackage{array,makecell}
\usepackage{arydshln}
\usepackage{textcomp}
\usepackage{ifthen}
\usepackage{graphicx}
\usepackage{xspace}
\usepackage{amsmath}
\usepackage{algorithm}
\usepackage{algpseudocode}
\usepackage{wrapfig}
\usepackage{subcaption}

\usepackage{listings}

\newcommand{\name}{VeriGraph\xspace}

\newcommand{\Sref}[1]{Section~\ref{#1}}

\newcommand{\Fref}[1]{Figure~\ref{#1}}
\newcommand{\Tref}[1]{Table~\ref{#1}}
\newcommand{\Aref}[1]{Algorithm~\ref{#1}}

\title{\bf
VeriGraph: Scene Graphs for Execution Verifiable Robot Planning
}

\author{Daniel Ekpo$^{1}$, Mara Levy$^{1}$, Saksham Suri$^{1}$, Chuong Huynh$^{1}$, Archana Swaminathan$^{1}$, Abhinav Shrivastava$^{1}$
\thanks{$^{1}$University of Maryland, College Park, MD USA.
        Correspondence to: Daniel Ekpo {\tt\small daniekpo@umd.edu}}%
}

\makeatletter
\let\@oldmaketitle\@maketitle
\renewcommand{\@maketitle}{
  \@oldmaketitle
  \centering
  \includegraphics[width=\textwidth]{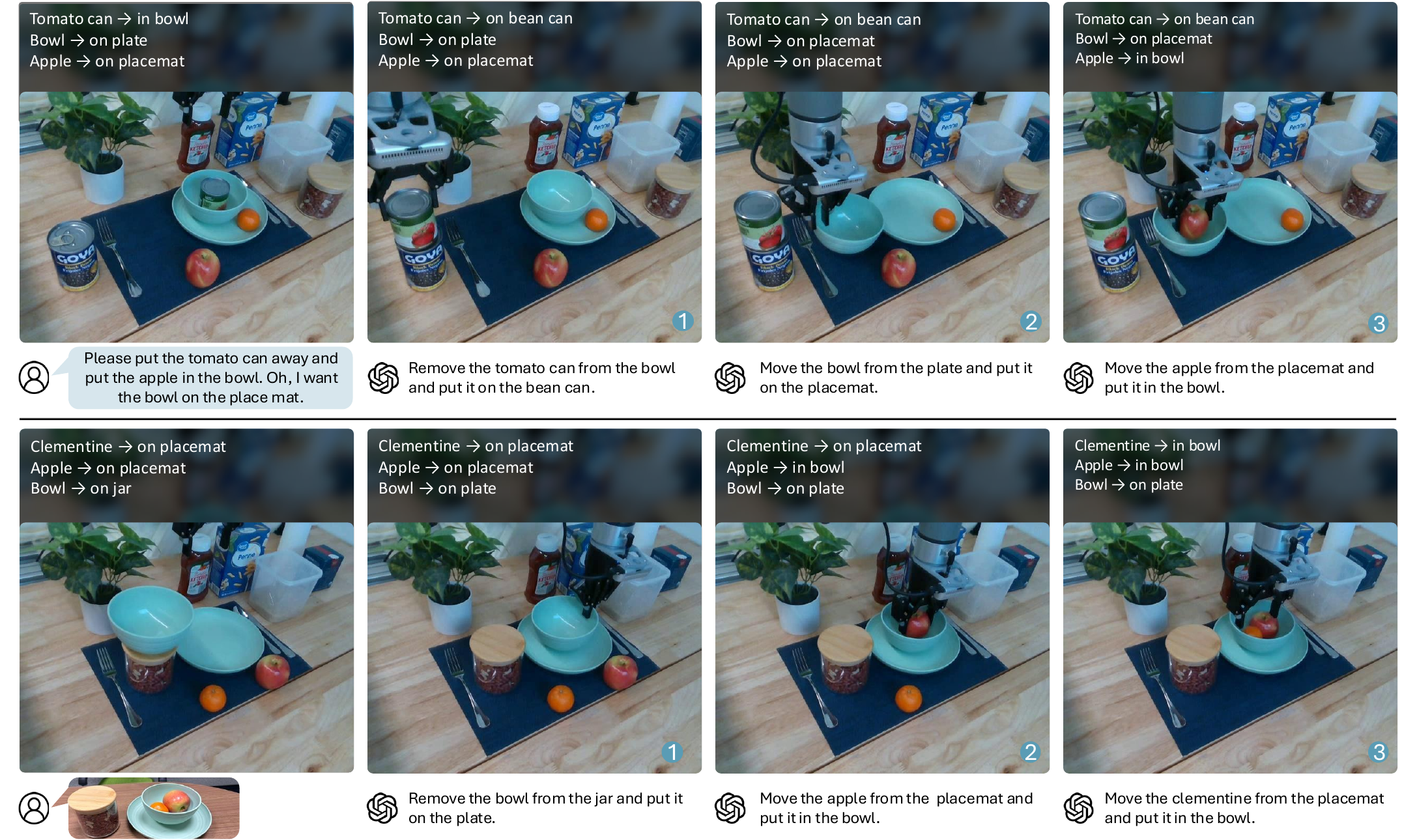}
  \captionof{figure}{VeriGraph leverages scene graphs and execution verification for long-horizon robot manipulation. Given an initial scene and a goal, specified as either a language instruction (top) or a reference image (bottom), a VLM planner generates and executes a sequence of actions, with verification after each step to ensure correctness before proceeding.}
  \vspace{-0.25in}
  \label{fig:teaser}
  \bigskip 
}
\makeatother

\begin{document}
\maketitle
\thispagestyle{empty}
\pagestyle{empty}

\begin{abstract}
Recent progress in vision-language models (VLMs) has opened new possibilities for robot task planning, but these models often produce incorrect action sequences. To address these limitations, we propose \name, a novel framework that integrates VLMs for robotic planning while verifying action feasibility. \name uses scene graphs as an intermediate representation to capture key objects and spatial relationships, enabling more reliable plan verification and refinement. The system generates a scene graph from input images and uses it to iteratively check and correct action sequences generated by an LLM-based task planner, ensuring constraints are respected and actions are executable. Our approach significantly enhances task completion rates across diverse manipulation scenarios, outperforming baseline methods by \(58\%\) on language-based tasks, \(56\%\) on tangram puzzle tasks, and \(30\%\) on image-based tasks. Qualitative results and code can be found at \url{https://verigraph-agent.github.io/}.
\looseness=-1

\end{abstract}

\section{Introduction}
\label{sec:intro}
For robots to be able to solve complex manipulation problems in the real world, they need to understand the physical world around them, including object locations and relationships between objects in the scene. Humans intuitively understand spatial relationships between objects in the world and can use this understanding to develop efficient and executable plans to complete tasks. Consider the example of organizing a cluttered room. Humans can quickly understand which objects are out of place based on their understanding of how objects relate to each other. For example, it seems intuitive that a book should be on a shelf, not on a cup. Robots struggle to perceive the world around them the way humans do.

Additionally, physical constraints in the real world restrict the order in which actions can be executed. For example, if a glass cup is on a book, which is on a desk, the robot must pick up the cup first and place it on the desk before picking up the book. Because of these constraints, robots need to understand the relationships between objects in the scene. If the robot does not understand that the cup is on the book, it might not factor that into its planning and may try to pick up the book first, which could result in the cup falling and breaking.

Recent advances in large language models (LLMs) and vision-language models (VLMs) have opened up new possibilities for robot task planning~\cite{huang2022zerolanguageplan,huang2023groundedrobot}. These models demonstrate impressive reasoning capabilities and world knowledge. Prior work~\cite{simon2021natural, wang2024grammar, Oswald_Srinivas_Kokel_Lee_Katz_Sohrabi_2024} used LLMs to generate Planning Domain Definition Language (PDDL), which can be used by classical planners to create a task plan. While the results have been promising, PDDL is inherently restrictive and does not generalize well~\cite{espasa2023challengesmodellingsolvingplotting, zhang2024dkpromptdomainknowledgeprompting}. Other lines of work use VLMs to generate high-level task plans directly from images~\cite{saycan, VILA, li2023interactive}, treating visual observation as the primary input to the planner.

Although these methods show promising results, long-horizon manipulation tasks require reasoning about objects, spatial relations, and physical constraints. Planning directly from raw images can therefore be challenging. While pixel-level representations contain rich visual detail, they can also introduce noise and irrelevant visual variation that complicate high-level reasoning. Scene graphs provide a structured abstraction that explicitly represents objects and their relations, enabling reasoning about spatial constraints and action feasibility.

To address the scene representation challenge, we propose \name, a framework that uses scene graphs as an intermediate representation for robot task planning. Scene graphs have proven particularly valuable in robotic task planning~\cite{zhu2021hierarchical, gu2023conceptgraphs, amiri2022reasoning, rana2023sayplan}. Their structured nature enables the abstraction of object-level details into symbolic graphs, making them robust to noise while retaining essential information about object interactions. For example, a scene graph might encode that a "cup is on the table" or a "spoon is inside the cup," providing a framework for reasoning about actions such as moving the spoon or rearranging the objects in the scene. This abstraction is particularly useful for tasks where the precise visual appearance of objects is not critical. One such task is using a reference scene to arrange objects in another scene to look like the reference scene. Because of the scene graph representation, \name can solve this task significantly better than methods that rely on raw pixel data.

\begin{figure*}[t]
\centering
\includegraphics[width=\linewidth]{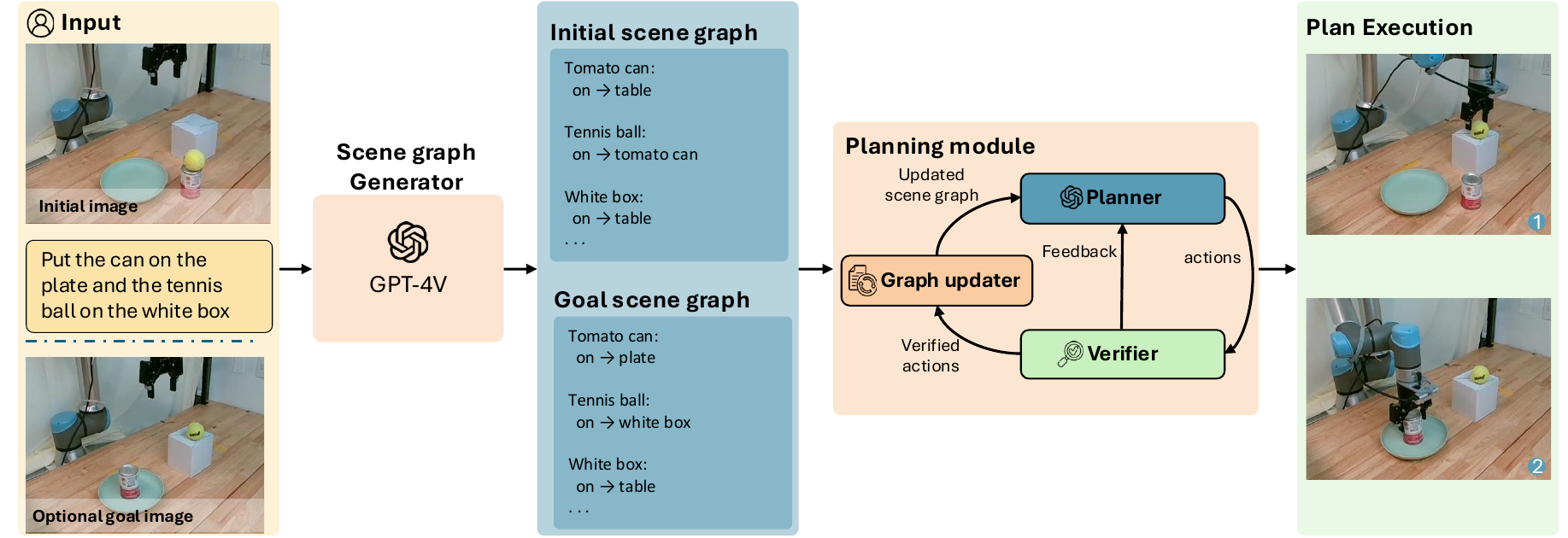}
\vspace{-0.2in}
  \caption{Overview of \name. Given a start image and either language instructions or a goal image, a scene-graph module extracts objects and relations to form the initial and goal scene graphs. \name then iteratively proposes high-level actions, validates them against graph-based constraints, and executes feasible actions to update the scene graph. On violation, \name generates feedback and replans; the loop ends when the planner emits an end token.}
\label{fig:approach}
\vspace{-0.07in}
\end{figure*}

Despite their strong capabilities, VLMs frequently produce incorrect plans and often require multiple prompting iterations to generate a valid action sequence~\cite{valmeekam2023planning}. Approaches such as~\cite{li2023interactive} attempt to solve this problem by inserting a human directly into the loop. However, this is time-consuming and requires constant human supervision. To address this plan verification problem, we add an iterative planning component to~\name. The structured nature of scene graphs allows \name to represent each action in the plan as graph operations. For example, moving an object from one location to another can be represented with an edge manipulation operation. This representation allows \name to quickly check for constraint violations and iterate with the task planner to generate valid action sequences. This setup, shown in Figure~\ref{fig:approach}, allows for more accurate and robust planning.\looseness=-1

Depending on the task, the goal state can be specified through either language instructions or a reference image. To support both types of task specification, \name supports flexible goal specification, allowing manipulation tasks to be defined through either target scene images or natural language instructions. \name can generate goal scene graphs from these input modalities, providing a unified planning framework across different task specifications. For reference images, \name does not require the same scene, only a contextually similar one (e.g., refer to~\Fref{fig:teaser}). This versatility makes \name applicable to various real-world scenarios where goals may be communicated in different formats.

We show that \name beats existing methods that rely on raw pixels as input to the task planner while being execution-verifiable. Our main contributions are as follows:
\begin{itemize}
    \item We present a modular framework that uses scene graphs as both the planning representation and the verification mechanism, enabling efficient constraint-aware planning with LLMs.
    \item We propose an iterative planning and verification mechanism that uses scene graphs to represent and verify action sequences, enhancing the system's ability to identify and correct constraint violations without human intervention.
    \item We utilize VLMs to generate goal scene graphs based on a reference image or language instruction to create a unified goal specification method.
\end{itemize}
\section{Related Work}
\label{sec:related}

\subsection{Scene Graphs in Planning}
Scene graphs have been widely used in computer vision for symbolic representation, capturing object relationships in images~\cite{krishna_visual_2016}. They have supported tasks such as image generation~\cite{johnson2018image, 9859841, mittal2019interactive, tripathi2019using, zhao2019image, yang2022diffusion}, image/video captioning~\cite{gao2018image,zhong2020comprehensive,yang2023transforming}, and visual question answering~\cite{zhang2019empirical,9031001,damodaran2021understanding}. By encoding high-level scene information while remaining robust to pixel-level noise, scene graphs have naturally motivated applications in robotics.

In robotic navigation and planning, scene graphs have been used to model environments and guide action selection. For example,~\cite{sepulvedadlindoorautonav} represent the environment as a graph of semantic places and navigational behaviors, while~\cite{ravichandr2022anmemory} leverage 3D scene graphs with graph neural networks (GNNs) to encode the scene and inform reinforcement learning policies. Other approaches integrate scene graphs with large models to enhance planning: GRID~\cite{ni2023grid} fuses an initial scene graph with a VLM encoder to guide actions, and~\cite{deng2023scene} uses object-detected scene graphs for cluttered scene exploration and task planning. Further extensions include augmenting only immobile objects~\cite{onishchenko2025lookplangraph}, multi-view scene graph generation with partially observable MDP solvers~\cite{amiri2022reasoning}, and combined geometric-symbolic graph representations for motion planning~\cite{zhu2021hierarchical}. SG-Bot~\cite{zhai2023sgbot} imagines goal scenes using scene graphs, and~\cite{jiao_sequential_2022} employ contact graphs with graph edit distance for sequential planning.

Our work builds on these graph-based planning approaches by modeling actions as graph operations. Unlike~\cite{jiao_sequential_2022}, we use an LLM as the task planner, allowing natural language task descriptions and greater generalization across objects, actions, and tasks, while still leveraging the structure of scene graphs for planning efficiency.

\subsection{Planning with LLMs/VLMs}
LLMs~\cite{touvron2023llama, jiang2023mistral,anil2023palm} and VLMs~\cite{openai2024gpt4,surís2023vipergpt,liu2023improved} have emerged as strong agents for open-world reasoning, understanding object relations, actions, and context. This has motivated their use as planners in robotics.

Prior work has explored using LLMs for high-level task planning~\cite{pramanick2020decomplex, venkatesh2021translating}, including verification of planned actions~\cite{grigorev2025verifyllmllmbasedpreexecutiontask} and replanning with visual feedback~\cite{pchelintsev2025lerareplanningvisualfeedback}. Embodied agents use LLMs to generate executable plans~\cite{saycan,huang2022inner,yao2022react}. While~\cite{huang2022inner} leverages language feedback, our method differs in using a scene graph as both the initial input and the continuous feedback to the planner, reducing data and computation requirements. SayCan~\cite{saycan} demonstrates strong results using learned value functions, but this approach requires substantial training data. ConceptGraphs~\cite{gu2023conceptgraphs} generate 3D scene graphs from 2D foundation models and plan with an LLM. In contrast, our approach operates directly on the scene graph representation and performs planning through graph-based constraint validation~(\Sref{sec:constraint_validation}) and iterative graph edits~(\Sref{sec:iterative_planning}), avoiding additional model fusion stages.\looseness=-1

Recent VLM developments have made them an even better candidate for robotics since they can take visual data and raw pixels as input, unlike LLMs, which can only work with text. This allows them to perceive the scene directly without being heavily dependent on the input prompt. Recently, RT-2~\cite{brohan2023rt} incorporated VLMs directly into end-to-end robotic control. PaLM-E~\cite{driess2023palm} trained a large model specifically as an agent that can take multimodal inputs and perform robotics tasks. VILA~\cite{VILA}, which is the closest to our work, uses a large VLM directly for their task. While these approaches show strong results, planning directly from pixels can be difficult for tasks that require reasoning about object relations and action constraints. Our approach instead introduces a structured scene graph representation that enables explicit constraint validation during planning.

\subsection{Execution-Verification}
While correctness is important when designing a task execution plan, it is also essential for it to be plausible. SayPlan~\cite{rana2023sayplan} assumes an existing 3D scene graph, which they use to interact with the LLM. They use a graph simulator to verify the LLM-generated task plans and show that their approach works for multi-room setups. CoPAL~\cite{joublin2023copal} proposes corrective planning by using different layers of encapsulation. Some works~\cite{mccallum2023feedback} modify an existing reinforcement learning algorithm to condition on natural language feedback from the environment. They automatically generate the language feedback based on the current goal and the agent's current actions. This is similar to our approach of providing feedback to the planner based on the current graph state and predicted action. REFLECT~\cite{liu2023reflect} introduces a framework to query the LLM planner to reason about failures based on the hierarchical summary of the robot's past experiences generated from multisensory observations. They show that the failure explanation can help the LLM correct the failure and complete the task. Voyager~\cite{wang2023voyager} introduces an LLM learner that can learn executable skills as it interacts with the environment. It writes code to interact with the environment and correct itself with feedback received from the environment. VILA~\cite{VILA} uses execution to verify the plan by feeding the current state of the environment to the model at every step using visual inputs. Our approach, on the other hand, can generate execution-verifiable plans by relying on the current scene graph for constraint checking. This makes verifying the affordances and plausibility of an action especially quick and efficient.


\section{Our Approach: VeriGraph}
\label{sec:approach}
\name takes in an image depicting an initial scene, along with either a target image portraying the desired goal scene state or instructions detailing modifications to the initial scene. It generates the initial and goal scene graphs and uses them to predict actions to transform the initial scene into the target scene. The scene graph generation method is discussed in~\Sref{sec:sg_gen}, and action constraints are discussed in~\Sref{sec:constraint_validation}. Given a pair of scene graphs, the planner, discussed in~\Sref{sec:task_planning} and~\Sref{sec:iterative_planning}, generates a high-level plan instructing a robot to transform the initial scene into the goal scene.

\subsection{Scene Graph Generation}
\label{sec:sg_gen}
Given an image $I$ of a scene, the goal of scene graph generation is to create a graph that accurately represents the scene's structure. The scene graph comprises a set of vertices $V$, representing objects in the scene, and a set of edges $E$, describing the relationships between objects in $V$. $R$ represents a set of possible relations between objects in the scene. An edge $e_{uv} \in E$ between two vertices $u$ and $v$ in the scene is then defined as $e_{uv} = \{u, v, r\}$ where $r \in R$ and $u, v \in V$. The scene graph for image $I$ is thus represented as $G = \{V, E\}$.

In \name, the set of relations $R$ includes basic spatial relations such as \{\text{in}, \text{on}\}. However, this set is flexible and easily adapted for other tasks. To address the issue of varying object names (e.g., tabletop, table, and countertop for the same object), we maintain a global dictionary of unique object names $D$ for scene graph generation. This dictionary encompasses all objects that could be present in any scene. The scene graph generator $SGG$ takes the image $I$, the global dictionary $D$, the set of relations $R$, and the task description $T$ and returns a scene graph. We define the graph generator as:\looseness=-1
\begin{equation}
    SGG(I, R, D, T) \rightarrow G = \{V, E\}
    \label{eq:sgg}
\end{equation}

When generating scene graphs, $T$ is set to null except for the target scene graph for tasks where the target scene is described using natural language, in which case the task instruction/goal scene description $T$ is used in the graph generation process. \name uses $SGG$ to generate the initial and goal scene graphs. 

\subsection{Constraint Validation}
\label{sec:constraint_validation}
Every action has a set of preconditions that must be met before execution. For example, before performing the "move" action on a plate, any items on the plate must first be removed. Additionally, post-conditions must be satisfied for the action to be considered successful. In the example given, the plate must end up on the new supporting object. In \name, these conditions are represented as a set of constraints $\mathcal{C}$.

\name uses the current graph to validate constraints. The node $v$ associated with the action must exist in the graph, and its in/out edges must satisfy specific conditions. For instance, if $v$ is being moved, \name checks if $v$ supports any other objects by ensuring no edges from $v$ to any other nodes exist.

After constraints are validated, \name updates the current graph state to reflect the execution of the action. The specific changes to the graph depend on the action taken. For a "move" operation, the edge representing the initial support relationship is removed, and a new edge is created for the new support relationship. The next action in the sequence is then validated, and the graph is modified accordingly. The final graph's nodes and edges are compared against the goal scene graph. If they match, the plan is considered successful.

\begin{figure}[t]
\centering
\includegraphics[width=\linewidth]{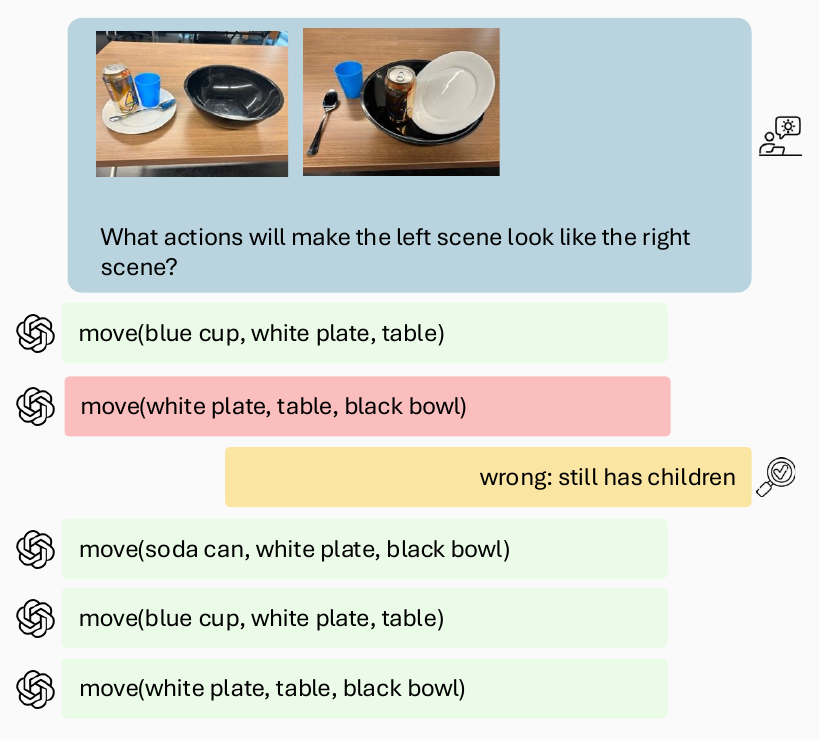}
\vspace{-0.35in}
  \caption{Iterative planning with graph-based feedback. \name proposes an action, validates it on the current graph, and executes it if feasible (green). When an action violates constraints (red), \name generates feedback and replans. This repeats until the planner signals completion.}
\label{fig:action_example}
\vspace{-0.2in}
\end{figure}

\subsection{Task Planning}
\label{sec:task_planning}
Given the initial scene graph $G_i$ and the target scene graph $G_g$, the task planner $\mathcal{P}$ generates actions that can be executed on the initial scene to transform it into the target graph. Let $\mathcal{A}$ be the set of all high-level actions that a robot can perform, the sequence of actions $a=\{a_1, a_2,\cdots,a_n\}$ such that $a \in \mathcal{A}$ predicted by the planner $\mathcal{P}$ must complete the given task while adhering to the constraints $\mathcal{C}$. We define the task planner as
\begin{equation}
    a=\mathcal{P}(G_i, G_g, \mathcal{C}, \mathcal{A}).
    \label{eq:planner}
\end{equation}
An example of such a plan can be seen in Figure~\ref{fig:action_example}.

\subsection{Iterative Planning}
\label{sec:iterative_planning}
The planner described in~\Sref{sec:task_planning} outputs the full action sequence without any feedback mechanism to refine the plan. Our experiments showed that this approach often fails on difficult tasks because LLMs tend to forget constraints. To address this issue, we designed an iterative planner $\mathcal{P}_{\text{iter}}$ that receives feedback $\mathcal{F}$ about the proposed action sequences and corrects the plan accordingly.

The planner $\mathcal{P}_\text{iter}$ is given $\mathcal{A}$, $\mathcal{C}$, $G_i$, and $G_g$ and asked to output at most $k$ high-level actions and an end token. \name attempts to perform the actions, and \name returns feedback, $\mathcal{F}$, as well as the current graph state. If there is a constraint violation in the proposed actions, the error count $\tau$ is increased. The feedback and new graph state are then passed to $\mathcal{P}_\text{iter}$, and the iteration continues until either the number of errors reaches the error threshold $\mathbf{t}$ or the end token is received. Because the LLM planner can repeatedly propose actions that violate the same constraints, the process may enter an error cycle. The error threshold prevents infinite replanning by limiting the number of failed iterations. The iterative planning process is described in~\Aref{alg:iterative}.

\begin{algorithm}[t]
\caption{Iterative Planning Algorithm}
\label{alg:iterative}
\begin{algorithmic}[1]
\Require Initial graph $G_i$, goal graph $G_g$, constraints $C$, action set $A$, actions per iteration $k$, error threshold $t$
\State $G \leftarrow G_i$, $F \leftarrow \emptyset$, $\tau \leftarrow 0$
\State $\hat{a} \leftarrow [\,]$ \Comment{executed actions}
\While{$\tau < t$}
    \State $(a_{1:k}, \texttt{end\_token}) \leftarrow P_{\text{iter}}(G, G_g, C, A, F)$
    \If{\texttt{end\_token}}
        \State \textbf{break}
    \EndIf
    \State $(\tilde{a},\, G,\, F) \leftarrow \text{execute\_until\_violation}(a_{1:k}, G, C)$
    \State append $\tilde{a}$ to $\hat{a}$
    \If{$F \neq \emptyset$} \Comment{constraint violation}
        \State $\tau \leftarrow \tau + 1$
    \Else
        \State $\tau \leftarrow 0$
    \EndIf
\EndWhile
\State \Return ($\hat{a}$, $G$)
\end{algorithmic}
\end{algorithm}

\section{Experimental Details}
\label{sec:experiments}

In this section, we outline the details of our experiments. The evaluation dataset is introduced in~\Sref{sec:exp_data}, followed by the task description in~\Sref{sec:exp_task}. All baselines are mentioned in~\Sref{sec:exp_baseline} and finally~\Sref{sec:exp_result} discusses the results.

\begin{figure}[t]
    \centering
    \includegraphics[width=\linewidth]{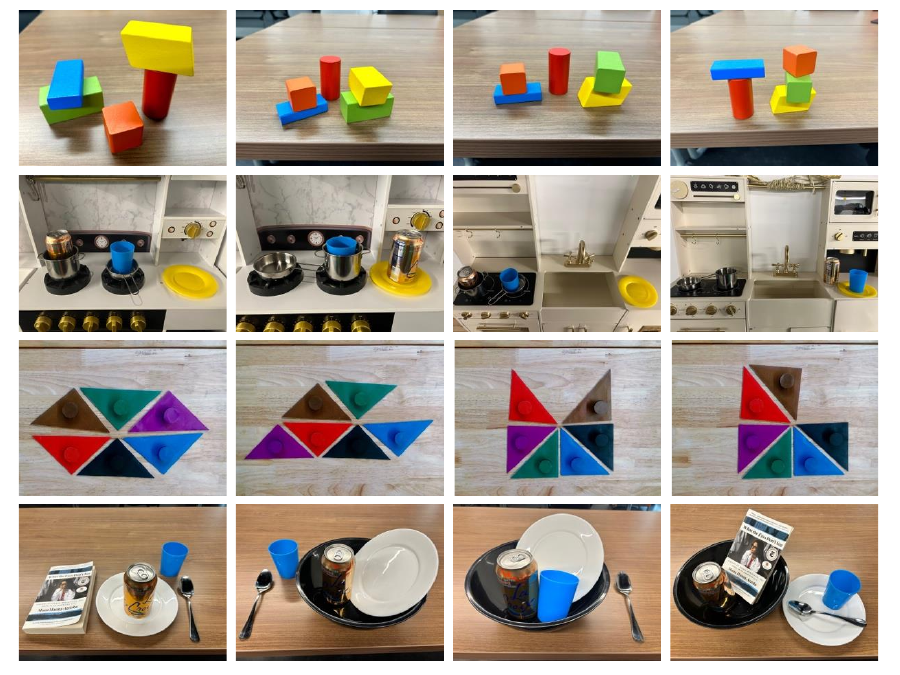}
    \vspace{-0.25in}
    \caption{Example evaluation scenes from the four task groups (blocks, kitchen, tangram, and tableware), illustrating varied object sets and support relations used to test constraint-aware planning.}
    \label{fig:dataset_examples}
    \vspace{-0.25in}
\end{figure}

\subsection{Dataset}
\label{sec:exp_data}
We design three scenes---kitchen, tableware, and block scenes---to evaluate \name's performance and compare against baseline methods. Each scene has multiple configurations with varying numbers of objects and placements. We vary the number of objects between three and seven. Ground truth scene graphs for each scene were created using GPT-4V(ision) (GPT-4V)~\cite{openai2023gpt4vision}, and corrections were made manually when necessary. Figure~\ref{fig:dataset_examples} shows some example images from the dataset.\looseness=-1

\subsection{Tasks}
\label{sec:exp_task}
We created four task groups - tangram puzzle, rearrange, language instruction, and stacking - with varying difficulty levels. Each task has a ground truth goal scene graph for evaluating the planner. The predicted actions from the planner are executed on the initial scene graph, and the transformed graph is then compared against the ground truth goal scene graph. The different task groups are described below.

\noindent\textbf{Tangram puzzle.} The task is defined by two images: a goal tangram configuration and a perturbed one. The goal can be reached by moving a single piece so that one of its sides connects with a specific side of another piece. The model must predict which piece to move, where to place it, and which sides should align. We detect tangrams with a triangle detector and use OpenClip~\cite{openclip_ilharco_gabriel_2021} to assign consistent color-based labels; sides are indexed clockwise. Contacts are computed using a heuristic module. In \textsc{VILA}, we provide labeled images, while in \name each configuration is represented as a scene graph with nodes as tangram pieces and edges as side-to-side contacts.

\noindent\textbf{Stacking Task.} We use block scenes of different configurations for the stacking task. The model is given an image of the initial scene and asked to stack all the blocks into one stack. The final order is arbitrary here, so there's no ground truth scene graph. Instead, the final scene graph is checked to ensure a single pile of blocks. Some block scenes already have multiple incomplete stacks, so the planner must unstack the other stacks and complete one.

\noindent\textbf{Language Instruction Task.} This consists of an initial scene and a language instruction. The instruction is either direct commands, e.g., "Move pan to the stovetop," or a description of the desired goal state of the scene, e.g., "I need the positions of the pan and pot swapped." The model is asked to predict actions to execute the given instruction on the scene.

\noindent\textbf{Reference Image Instruction Task.} The model is given an initial scene and a structurally similar scene as the goal state and asked to predict a sequence of actions to transform the initial scene into the goal scene. The scene graph of the goal scene is compared against the final scene graph after plan execution, and the task is considered successful if the scene graphs match.

\begin{table}[t]
   \setlength{\abovecaptionskip}{7pt}
   \caption{Results on scene graph generation.}
   \label{tab:sg_blocks}
   \centering
   \setlength{\cmidrulewidth}{0.01em}
   \renewcommand{\tabcolsep}{10pt}
   \renewcommand{\arraystretch}{1.2}
   \begin{tabular}{@{}llccc@{}}
   \toprule
        \multirow{2}{*}{Scene} & \multirow{2}{*}{Method} & \multicolumn{2}{@{}c@{}}{F1 Score} & \multirow{2}{*}{Exact Graph Accuracy}\\
        \cmidrule[\cmidrulewidth](lr){3-4}

          & & Nodes & Edges &\\
         \midrule
         \multirow{3}{*}{Blocks} &LLaVA & 0.30 & 0.07 & 0.00 \\
         &Gemini & \textbf{1.00} & 0.93 & 0.73 \\
         & GPT-4V & \textbf{1.00} & \textbf{1.00} & \textbf{1.00} \\
         \hdashline\noalign{\vskip 0.5ex}

         \multirow{3}{*}{Kitchen} &LLaVA & 0.50 & 0.05 & 0.00 \\

         &Gemini & 0.78 & 0.59 & 0.04 \\
         & GPT-4V & \textbf{0.98} & \textbf{0.87} & \textbf{0.65} \\
         \hdashline\noalign{\vskip 0.5ex}
         \multirow{3}{*}{Tableware}&LLaVA & 0.52 & 0.16 & 0.00 \\
         &Gemini & 0.95 & 0.90 & 0.39 \\
         & GPT-4V & \textbf{0.99} & \textbf{0.95} & \textbf{0.89} \\
         \bottomrule
   \end{tabular}
\end{table}

\subsection{Baselines}
\label{sec:exp_baseline}
We compare \name against the following baselines: (a) \textbf{VILA}~\cite{VILA} prompts the vision-language model (VLM) with an image and a language instruction or another image as the goal state. For a fair comparison, we use a prompt identical to ours and remove all references to scene graphs. We execute the proposed actions on the ground truth initial scene graph and compare the final graph against the ground truth graph. (b) \textbf{SayCan}~\cite{saycan} prompts the LLM with a textual representation of the scene. The text contains a list of all the objects in the scene. We implemented a similar setup using GPT-4 as the LLM. Since SayCan cannot understand the spatial relationships between objects in the scene because it only receives a list of all objects, we only evaluate it on image-language tasks.
The objects from the vertices in the ground truth scene graph are the scene observation for this baseline.

\subsection{Experiments and Results}
\label{sec:exp_result}
\textbf{Scene graph generation:} As mentioned earlier, VLMs are utilized in our scene graph generation model. Here, we evaluate some popular VLMs, such as Gemini 1.5 Pro~\cite{googlegeminifamilyhighlycapable}, LLaVA~\cite{liu2023llava}, and GPT-4V~\cite{openai2023gpt4vision}, without any in-context examples. For GPT-4V and Gemini 1.5 Pro, we use the official Python SDK. Ollama~\cite{githubollama} is used to run LLaVA locally. The same prompt is used for all three models. For our experiments, the set of relations $R$ and global dictionary $D$ are predefined and included in the prompt. We evaluate node and edge prediction using F1-score compared to the ground truth scene graph. We also report Exact Graph Accuracy, which requires a perfect match of all nodes and edges in the predicted graph. Because this metric requires the entire graph to be correct simultaneously, it is substantially stricter than the per-element F1 scores. The three models are compared across the scenes in ~\Tref{tab:sg_blocks}. GPT-4V performs notably better than Gemini and LLaVA across all scenes. We used GPT-4V as the scene graph generator $SGG$ for other experiments based on these results.

\begin{table}
  \setlength{\abovecaptionskip}{7pt}
  \caption{Planning results across different task types. We report task success rate, where success indicates that the final scene graph after executing the predicted actions matches the goal scene graph. ``Ours (Direct)'' uses a single-shot planner without verification, while ``Ours'' uses the iterative planning and constraint-validation mechanism of VeriGraph, which corrects actions that violate scene graph constraints.}
  \label{tab:planning_results}
  \centering
  \setlength{\cmidrulewidth}{0.01em}
  \renewcommand{\tabcolsep}{4pt}
  \renewcommand{\arraystretch}{1.1}
  \begin{tabular}{@{}lcccc@{}}
    \toprule
         Task & SayCan & VILA & Ours (Direct) & Ours \\
         \midrule
         Stacking & 0.07&0.62&0.35&\textbf{0.65}\\
         Language Instruction & 0.17&0.43&\textbf{0.73}&0.65\\
         Ref.\ Image Instruction (Blocks) & -&0.27&0.67&\textbf{0.86}\\
         Ref.\ Image Instruction  (Kitchen) & 0.00 &0.05&0.50&\textbf{0.55}\\
         Tangram puzzle & - & 0.16 & 0.72 & \textbf{0.72} \\
         \bottomrule
    \end{tabular}
\end{table}

\textbf{Planning:} We used GPT-4~\cite{openai2024gpt4} as the task planner $\mathcal{P}$. We evaluate our approach on all four task groups presented in~\Sref{sec:exp_task} and show results in~\Tref{tab:planning_results}. Compared to VILA, we improve on language and image instruction planning tasks. For the image-based start and goal state, we observe an average improvement of ${\sim}0.57$. For the language instruction task, we outperform SayCan by ${\sim}0.56$. Similarly, for the tangram puzzle tasks we outperform VILA by \(56\%\).

We attribute the performance improvements to the structured representation of the scene as a scene graph. Scene graph representations structure the environment into explicit objects and relations, reducing the need for the VLM to infer spatial relationships directly from raw images and enabling iterative correction during the planning stage. Most failure cases in our proposed approach arise from inaccuracies in the generated scene graphs. As scene graph generation methods improve, the effectiveness of our planning approach is expected to improve as well. To evaluate the impact of scene graph quality, we provided the planner with ground truth scene graphs and observed that the iterative planner generated successful plans in nearly all cases.

Our results in~\Tref{tab:planning_results} show that the iterative planner (last column) improves performance for most tasks compared to the non-iterative planner. However, the language instruction task shows a slight decrease and the tangram task shows no change. One contributing factor is the accuracy of the generated scene graphs. The iterative planner relies on graph constraints to detect invalid actions, and errors in the scene graph can prevent these violations from being detected correctly. This is consistent with our earlier observation that when ground truth scene graphs are provided, the iterative planner produces successful plans in nearly all cases.

\begin{table}[!t]
  \caption{Ablation on error threshold and number of actions per iteration used in \name.}
  \vspace{-0.05in}
  \begin{subtable}[t]{0.48\linewidth}
    \renewcommand{\tabcolsep}{2pt}
    \renewcommand{\arraystretch}{1.1}
    \centering
    \caption{Effect of error threshold.}
    \label{tab:error_t}
    \vspace{-0.05in}
    \resizebox{\linewidth}{!}{%
      \begin{tabular}{@{}lcccc@{}}
        \toprule
        Error Threshold ($\tau$) & 2 & 3 & 5& 10 \\
        \midrule
        Accuracy (\%) & 20 & 85 & 80 & 90 \\
        \bottomrule
      \end{tabular}
    }
  \end{subtable}
  \hfill
  \begin{subtable}[t]{0.48\linewidth}
    \renewcommand{\tabcolsep}{4pt}
    \renewcommand{\arraystretch}{1.1}
    \centering
    \caption{Varying the number of actions per iteration.}
    \label{tab:iterations}
    \vspace{-0.05in}
    \resizebox{\linewidth}{!}{%
      \begin{tabular}{@{}lcccc@{}}
        \toprule
        \# of Actions & 2 & 3 & 5 & 10 \\
        \midrule
        Accuracy (\%) & 85 & 85 & 85 & 85\\
        \bottomrule
      \end{tabular}
    }
  \end{subtable}%
\vspace{-0.2in}
\end{table}

\subsection{Ablation Study}
\label{sec:exp_abl}
\textbf{Iterative vs.\ non-iterative planner.} Our experiment results in~\Tref{tab:planning_results} show that the iterative planner ('Ours') achieves higher accuracy than the non-iterative planner ('Ours (Direct)') for most tasks. The feedback in the iterative process allows for iterative correction during the planning stage, which results in improved task planning accuracy.

\textbf{Error thresholds.} We observed that setting the error threshold to 2 resulted in the worst performance. While setting it to 10 yielded the best performance, it increased runtime and cost due to additional replanning iterations. We found that setting the threshold to 5 provided a good balance between accuracy and speed.

\textbf{Number of actions per iteration.} We tested 2, 3, 5, and 10 actions per iteration. Although there was no significant difference in accuracy, we ultimately chose to use 3 actions per iteration for optimal performance.

\subsection{Real-world execution}
\label{sec:real_world}
To demonstrate that the high-level plans generated by VeriGraph can be executed on a physical robot, we deploy the system on representative manipulation tasks described in~\Sref{sec:exp_task}. We implement a modular execution pipeline that converts high-level task plans to low-level robot commands. We use LangSAM~\cite{medeiros2023langsam} to obtain object masks for the target and destination objects. The detected objects are projected to 3D coordinates and transformed to the robot frame. We use AnyGrasp~\cite{fang2023anygrasp} to predict a grasp pose for the target object, and estimate the drop height from the point cloud of the destination object.

For the tangram task, we compute the x–y offset between the centers of the target and destination pieces, along with the rotation needed to align the specified sides. These offsets are then projected into 3D to produce the final robot commands.
The generated plans are executed in an open-loop manner. Qualitative results of the real-world executions are available on the project website.

\subsection{Cost and runtime analysis}
\label{sec:cost_analysis}
The computational cost of VeriGraph is dominated by calls to the VLM used for scene graph generation and the LLM used for task planning. Two VLM calls are required to generate the initial and goal scene graphs. The monetary cost therefore depends on the specific model provider; since these costs change frequently and continue to decrease, we do not report fixed numbers here.

The cost of planning depends on the number of iterations and the number of actions proposed per iteration. Proposing more actions per iteration can reduce the number of LLM calls, lowering overall planning cost. In addition to monetary cost, system latency is primarily determined by network bandwidth and the response time of the models.

The graph-based constraint validation and state updates are lightweight operations whose runtime is negligible compared to model inference. This enables flexible deployment strategies. In particular, our experiments suggest that while local models often struggle with accurate scene graph generation, they can produce effective plans when provided with a correct scene graph and combined with the iterative verification mechanism. This suggests a hybrid approach in which a high-capability VLM generates the scene graph, while a local LLM performs task planning, significantly reducing cost while maintaining strong performance.
\section{Conclusion}
In this paper, we presented \name, a framework for generating and validating high-level task plans for robot object manipulation using scene graphs. \name represents actions as graph edit operations and verifies them using node- and edge-based constraints, enabling efficient validation of planner outputs before execution. We demonstrate that scene graphs provide a compact and structured representation of scene information and show that our method outperforms approaches that rely directly on raw pixel inputs. We also introduced an iterative planning mechanism in which proposed actions are validated against the current graph state and corrected when constraint violations occur, improving the robustness of the planning process. In addition, the framework provides an efficient way to evaluate high-level robot task-planning algorithms.

Our approach depends on the accuracy of the generated scene graphs. Because scene graphs are extracted from monocular images, errors in object detection or spatial relations can propagate to the planning stage and lead to incorrect actions. In practice, we observed that hallucinated objects or incorrect relations occasionally produced invalid plans. While the constraint validation stage prevents some infeasible actions, it cannot correct structural errors in the underlying graph. Improving the reliability of scene graph generation therefore remains an important direction for future work. As scene graph generation methods continue to improve, the planning performance of \name is expected to improve correspondingly.

\section{Acknowledgement}
This work was partially supported by NSF CAREER Award \(\#2238769\) to AS. The U.S. Government is authorized to reproduce and distribute reprints for Governmental purposes notwithstanding any copyright annotation thereon. The views and conclusions contained herein are those of the authors and should not be interpreted as necessarily representing the official policies or endorsements, either expressed or implied, of NSF or the U.S. Government. We thank Namitha Padmanabhan, Pulkit Kumar, Seungjae Lee, and Vatsal Agarwal for their helpful feedback and discussions.



\bibliographystyle{IEEEtran}
\bibliography{root}

@String(CVPR  = {IEEE Conf. Comput. Vis. Pattern Recog.})

@String(ECCV  = {Eur. Conf. Comput. Vis.})

@String(NeurIPS = {Adv. Neural Inform. Process. Syst.})

@String(ICML  = {Int. Conf. Mach. Learn.})

@String(ICME  = {Int. Conf. Multimedia and Expo})

@String(CORL  = {Conference on Robot Learning})

@String(ICAPS = {Int. Conf. Autom. Planning Scheduling})

@String(ICRA  = {IEEE Int. Conf. Robot. Autom.})

@String(IROS  = {IEEE/RSJ Int. Conf. Intell. Robots Syst.})

@String(TRO   = {IEEE Trans. Robot.})

@article{vila,
  title   = {Look before you leap: Unveiling the power of gpt-4v in robotic vision-language planning},
  author  = {Hu, Yingdong and Lin, Fanqi and Zhang, Tong and Yi, Li and Gao, Yang},
  journal = {arXiv preprint arXiv:2311.17842},
  year    = {2023}
}

@misc{saycan,
  title         = {Do As I Can, Not As I Say: Grounding Language in Robotic Affordances},
  author        = {Michael Ahn et. al.},
  year          = {2022},
  eprint        = {2204.01691},
  archiveprefix = {arXiv},
  primaryclass  = {cs.RO}
}

@misc{huang2022inner,
  title         = {Inner Monologue: Embodied Reasoning through Planning with Language Models},
  author        = {Wenlong Huang and Fei Xia and Ted Xiao and Harris Chan and Jacky Liang and Pete Florence and Andy Zeng and Jonathan Tompson and Igor Mordatch and Yevgen Chebotar and Pierre Sermanet and Noah Brown and Tomas Jackson and Linda Luu and Sergey Levine and Karol Hausman and Brian Ichter},
  year          = {2022},
  eprint        = {2207.05608},
  archiveprefix = {arXiv},
  primaryclass  = {cs.RO}
}

@misc{surís2023vipergpt,
  title         = {ViperGPT: Visual Inference via Python Execution for Reasoning},
  author        = {Dídac Surís and Sachit Menon and Carl Vondrick},
  year          = {2023},
  eprint        = {2303.08128},
  archiveprefix = {arXiv},
  primaryclass  = {cs.CV}
}

@misc{li2023interactive,
  title         = {Interactive Task Planning with Language Models},
  author        = {Boyi Li and Philipp Wu and Pieter Abbeel and Jitendra Malik},
  year          = {2023},
  eprint        = {2310.10645},
  archiveprefix = {arXiv},
  primaryclass  = {cs.RO}
}

@misc{openai2024gpt4,
  title         = {GPT-4 Technical Report},
  author        = {OpenAI},
  year          = {2024},
  eprint        = {2303.08774},
  archiveprefix = {arXiv},
  primaryclass  = {cs.CL}
}

@article{driess2023palm,
  title   = {Palm-e: An embodied multimodal language model},
  author  = {Driess, Danny and Xia, Fei and Sajjadi, Mehdi SM and Lynch, Corey and Chowdhery, Aakanksha and Ichter, Brian and Wahid, Ayzaan and Tompson, Jonathan and Vuong, Quan and Yu, Tianhe and others},
  journal = {arXiv preprint arXiv:2303.03378},
  year    = {2023}
}

@article{brohan2023rt,
  title   = {Rt-2: Vision-language-action models transfer web knowledge to robotic control},
  author  = {Brohan, Anthony and Brown, Noah and Carbajal, Justice and Chebotar, Yevgen and Chen, Xi and Choromanski, Krzysztof and Ding, Tianli and Driess, Danny and Dubey, Avinava and Finn, Chelsea and others},
  journal = {arXiv preprint arXiv:2307.15818},
  year    = {2023}
}

@article{liu2023improved,
  title   = {Improved baselines with visual instruction tuning},
  author  = {Liu, Haotian and Li, Chunyuan and Li, Yuheng and Lee, Yong Jae},
  journal = {arXiv preprint arXiv:2310.03744},
  year    = {2023}
}

@article{anil2023palm,
  title   = {Palm 2 technical report},
  author  = {Anil, Rohan and Dai, Andrew M and Firat, Orhan and Johnson, Melvin and Lepikhin, Dmitry and Passos, Alexandre and Shakeri, Siamak and Taropa, Emanuel and Bailey, Paige and Chen, Zhifeng and others},
  journal = {arXiv preprint arXiv:2305.10403},
  year    = {2023}
}

@article{jiang2023mistral,
  title   = {Mistral 7B},
  author  = {Jiang, Albert Q and Sablayrolles, Alexandre and Mensch, Arthur and Bamford, Chris and Chaplot, Devendra Singh and Casas, Diego de las and Bressand, Florian and Lengyel, Gianna and Lample, Guillaume and Saulnier, Lucile and others},
  journal = {arXiv preprint arXiv:2310.06825},
  year    = {2023}
}

@article{touvron2023llama,
  title   = {Llama: Open and efficient foundation language models},
  author  = {Touvron, Hugo and Lavril, Thibaut and Izacard, Gautier and Martinet, Xavier and Lachaux, Marie-Anne and Lacroix, Timoth{\'e}e and Rozi{\`e}re, Baptiste and Goyal, Naman and Hambro, Eric and Azhar, Faisal and others},
  journal = {arXiv preprint arXiv:2302.13971},
  year    = {2023}
}

@misc{zhai2023sgbot,
  title         = {SG-Bot: Object Rearrangement via Coarse-to-Fine Robotic Imagination on Scene Graphs},
  author        = {Guangyao Zhai and Xiaoni Cai and Dianye Huang and Yan Di and Fabian Manhardt and Federico Tombari and Nassir Navab and Benjamin Busam},
  year          = {2023},
  eprint        = {2309.12188},
  archiveprefix = {arXiv},
  primaryclass  = {cs.RO}
}

@misc{amiri2022reasoning,
  title         = {Reasoning with Scene Graphs for Robot Planning under Partial Observability},
  author        = {Saeid Amiri and Kishan Chandan and Shiqi Zhang},
  year          = {2022},
  eprint        = {2202.10432},
  archiveprefix = {arXiv},
  primaryclass  = {cs.RO}
}

@misc{zhu2021hierarchical,
  title         = {Hierarchical Planning for Long-Horizon Manipulation with Geometric and Symbolic Scene Graphs},
  author        = {Yifeng Zhu and Jonathan Tremblay and Stan Birchfield and Yuke Zhu},
  year          = {2021},
  eprint        = {2012.07277},
  archiveprefix = {arXiv},
  primaryclass  = {cs.RO}
}

@inproceedings{rana2023sayplan,
  title     = {SayPlan: Grounding Large Language Models using 3D Scene Graphs for Scalable Task Planning},
  author    = {Krishan Rana and Jesse Haviland and Sourav Garg and Jad Abou-Chakra and Ian Reid and Niko Suenderhauf},
  booktitle = CORL,
  year      = {2023}
}

@inproceedings{jiao_sequential_2022,
  address   = {Kyoto, Japan},
  author    = {Jiao, Ziyuan and Niu, Yida and Zhang, Zeyu and Zhu, Song-Chun and Zhu, Yixin and Liu, Hangxin},
  booktitle = IROS,
  doi       = {10.1109/IROS47612.2022.9981735},
  isbn      = {978-1-66547-927-1},
  month     = oct,
  pages     = {8203--8210},
  publisher = {IEEE},
  title     = {Sequential {Manipulation} {Planning} on {Scene} {Graph}},
  year      = {2022}
}

@misc{joublin2023copal,
  title         = {CoPAL: Corrective Planning of Robot Actions with Large Language Models},
  author        = {Frank Joublin and Antonello Ceravola and Pavel Smirnov and Felix Ocker and Joerg Deigmoeller and Anna Belardinelli and Chao Wang and Stephan Hasler and Daniel Tanneberg and Michael Gienger},
  year          = {2023},
  eprint        = {2310.07263},
  archiveprefix = {arXiv},
  primaryclass  = {cs.RO}
}

@misc{gu2023conceptgraphs,
  title         = {ConceptGraphs: Open-Vocabulary 3D Scene Graphs for Perception and Planning},
  author        = {Qiao Gu and Alihusein Kuwajerwala and Sacha Morin and Krishna Murthy Jatavallabhula and Bipasha Sen and Aditya Agarwal and Corban Rivera and William Paul and Kirsty Ellis and Rama Chellappa and Chuang Gan and Celso Miguel de Melo and Joshua B. Tenenbaum and Antonio Torralba and Florian Shkurti and Liam Paull},
  year          = {2023},
  eprint        = {2309.16650},
  archiveprefix = {arXiv},
  primaryclass  = {cs.RO}
}

@misc{deng2023scene,
  title         = {Scene Graph for Embodied Exploration in Cluttered Scenario},
  author        = {Yuhong Deng and Qie Sima and Di Guo and Huaping Liu and Yi Wang and Fuchun Sun},
  year          = {2023},
  eprint        = {2207.07870},
  archiveprefix = {arXiv},
  primaryclass  = {cs.RO}
}

@misc{ni2023grid,
  title         = {GRID: Scene-Graph-based Instruction-driven Robotic Task Planning},
  author        = {Zhe Ni and Xiao-Xin Deng and Cong Tai and Xin-Yue Zhu and Xiang Wu and Yong-Jin Liu and Long Zeng},
  year          = {2023},
  eprint        = {2309.07726},
  archiveprefix = {arXiv},
  primaryclass  = {cs.RO}
}

@inproceedings{ravichandr2022anmemory,
  author    = {Ravichandran, Zachary and Peng, Lisa and Hughes, Nathan and Griffith, J. Daniel and Carlone, Luca},
  booktitle = ICRA,
  title     = {Hierarchical Representations and Explicit Memory: Learning Effective Navigation Policies on 3D Scene Graphs using Graph Neural Networks},
  year      = {2022},
  volume    = {},
  number    = {},
  pages     = {9272-9279},
  doi       = {10.1109/ICRA46639.2022.9812179}
}

@misc{krishna_visual_2016,
  title      = {Visual {Genome}: {Connecting} {Language} and {Vision} {Using} {Crowdsourced} {Dense} {Image} {Annotations}},
  shorttitle = {Visual {Genome}},
  publisher  = {arXiv},
  author     = {Krishna, Ranjay and Zhu, Yuke and Groth, Oliver and Johnson, Justin and Hata, Kenji and Kravitz, Joshua and Chen, Stephanie and Kalantidis, Yannis and Li, Li-Jia and Shamma, David A. and Bernstein, Michael S. and Li, Fei-Fei},
  month      = feb,
  year       = {2016}
}

@inproceedings{johnson2018image,
  author    = {Johnson, Justin and Gupta, Agrim and Fei-Fei, Li},
  booktitle = CVPR,
  title     = {Image Generation from Scene Graphs},
  year      = {2018},
  volume    = {},
  number    = {},
  pages     = {1219-1228},
  doi       = {10.1109/CVPR.2018.00133}
}

@article{zhang2019empirical,
  title   = {An empirical study on leveraging scene graphs for visual question answering},
  author  = {Zhang, Cheng and Chao, Wei-Lun and Xuan, Dong},
  journal = {arXiv preprint arXiv:1907.12133},
  year    = {2019}
}

@article{yang2023transforming,
  title   = {Transforming Visual Scene Graphs to Image Captions},
  author  = {Yang, Xu and Peng, Jiawei and Wang, Zihua and Xu, Haiyang and Ye, Qinghao and Li, Chenliang and Yan, Ming and Huang, Fei and Li, Zhangzikang and Zhang, Yu},
  journal = {arXiv preprint arXiv:2305.02177},
  year    = {2023}
}

@inproceedings{zhong2020comprehensive,
  title        = {Comprehensive image captioning via scene graph decomposition},
  author       = {Zhong, Yiwu and Wang, Liwei and Chen, Jianshu and Yu, Dong and Li, Yin},
  booktitle    = ECCV,
  pages        = {211--229},
  year         = {2020},
  organization = {Springer}
}

@inproceedings{gao2018image,
  title     = {Image captioning with scene-graph based semantic concepts},
  author    = {Gao, Lizhao and Wang, Bo and Wang, Wenmin},
  booktitle = {Proceedings of the 2018 10th international conference on machine learning and computing},
  pages     = {225--229},
  year      = {2018}
}

@article{yang2022diffusion,
  title   = {Diffusion-based scene graph to image generation with masked contrastive pre-training},
  author  = {Yang, Ling and Huang, Zhilin and Song, Yang and Hong, Shenda and Li, Guohao and Zhang, Wentao and Cui, Bin and Ghanem, Bernard and Yang, Ming-Hsuan},
  journal = {arXiv preprint arXiv:2211.11138},
  year    = {2022}
}

@inproceedings{zhao2019image,
  title     = {Image generation from layout},
  author    = {Zhao, Bo and Meng, Lili and Yin, Weidong and Sigal, Leonid},
  booktitle = CVPR,
  pages     = {8584--8593},
  year      = {2019}
}

@article{tripathi2019using,
  title   = {Using scene graph context to improve image generation},
  author  = {Tripathi, Subarna and Bhiwandiwalla, Anahita and Bastidas, Alexei and Tang, Hanlin},
  journal = {arXiv preprint arXiv:1901.03762},
  year    = {2019}
}

@article{mittal2019interactive,
  title   = {Interactive image generation using scene graphs},
  author  = {Mittal, Gaurav and Agrawal, Shubham and Agarwal, Anuva and Mehta, Sushant and Marwah, Tanya},
  journal = {arXiv preprint arXiv:1905.03743},
  year    = {2019}
}

@inproceedings{9859841,
  author    = {Zhao, Xin and Wu, Lei and Chen, Xu and Gong, Bin},
  booktitle = ICME,
  title     = {High-Quality Image Generation from Scene Graphs with Transformer},
  year      = {2022},
  volume    = {},
  number    = {},
  pages     = {1-6},
  doi       = {10.1109/ICME52920.2022.9859841}
}

@inproceedings{9031001,
  author    = {Lee, Soohyeong and Kim, Ju-Whan and Oh, Youngmin and Jeon, Joo Hyuk},
  booktitle = {2019 First International Conference on Graph Computing (GC)},
  title     = {Visual Question Answering over Scene Graph},
  year      = {2019},
  volume    = {},
  number    = {},
  pages     = {45-50},
  doi       = {10.1109/GC46384.2019.00015}
}

@article{damodaran2021understanding,
  title   = {Understanding the role of scene graphs in visual question answering},
  author  = {Damodaran, Vinay and Chakravarthy, Sharanya and Kumar, Akshay and Umapathy, Anjana and Mitamura, Teruko and Nakashima, Yuta and Garcia, Noa and Chu, Chenhui},
  journal = {arXiv preprint arXiv:2101.05479},
  year    = {2021}
}

@inproceedings{sepulvedadlindoorautonav,
  author    = {Sepulveda, G. and Niebles, J. C. and Soto, A.},
  booktitle = ICRA,
  title     = {A Deep Learning Based Behavioral Approach to Indoor Autonomous Navigation},
  year      = {2018},
  volume    = {},
  number    = {},
  pages     = {4646-4653},
  doi       = {10.1109/ICRA.2018.8460646}
}

@misc{mccallum2023feedback,
  title         = {Is Feedback All You Need? Leveraging Natural Language Feedback in Goal-Conditioned Reinforcement Learning},
  author        = {Sabrina McCallum and Max Taylor-Davies and Stefano V. Albrecht and Alessandro Suglia},
  year          = {2023},
  eprint        = {2312.04736},
  archiveprefix = {arXiv},
  primaryclass  = {cs.CL}
}

@misc{liu2023reflect,
  title         = {REFLECT: Summarizing Robot Experiences for Failure Explanation and Correction},
  author        = {Zeyi Liu and Arpit Bahety and Shuran Song},
  year          = {2023},
  eprint        = {2306.15724},
  archiveprefix = {arXiv},
  primaryclass  = {cs.RO}
}

@misc{wang2023voyager,
  title         = {Voyager: An Open-Ended Embodied Agent with Large Language Models},
  author        = {Guanzhi Wang and Yuqi Xie and Yunfan Jiang and Ajay Mandlekar and Chaowei Xiao and Yuke Zhu and Linxi Fan and Anima Anandkumar},
  year          = {2023},
  eprint        = {2305.16291},
  archiveprefix = {arXiv},
  primaryclass  = {cs.AI}
}

@article{yao2022react,
  title   = {React: Synergizing reasoning and acting in language models},
  author  = {Yao, Shunyu and Zhao, Jeffrey and Yu, Dian and Du, Nan and Shafran, Izhak and Narasimhan, Karthik and Cao, Yuan},
  journal = {arXiv preprint arXiv:2210.03629},
  year    = {2022}
}

@inproceedings{pramanick2020decomplex,
  title        = {DeComplex: Task planning from complex natural instructions by a collocating robot},
  author       = {Pramanick, Pradip and Barua, Hrishav Bakul and Sarkar, Chayan},
  booktitle    = {2020 IEEE/RSJ International Conference on Intelligent Robots and Systems (IROS)},
  pages        = {6894--6901},
  year         = {2020},
  organization = {IEEE}
}

@inproceedings{venkatesh2021translating,
  title        = {Translating Natural Language Instructions to Computer Programs for Robot Manipulation},
  author       = {Venkatesh, Sagar Gubbi and Upadrashta, Raviteja and Amrutur, Bharadwaj},
  booktitle    = {2021 IEEE/RSJ International Conference on Intelligent Robots and Systems (IROS)},
  pages        = {1919--1926},
  year         = {2021},
  organization = {IEEE}
}

@inproceedings{simon2021natural,
  title     = {A natural language model for generating pddl},
  author    = {Simon, Nisha and Muise, Christian},
  booktitle = {ICAPS KEPS workshop},
  year      = {2021}
}

@article{wang2024grammar,
  title   = {Grammar prompting for domain-specific language generation with large language models},
  author  = {Wang, Bailin and Wang, Zi and Wang, Xuezhi and Cao, Yuan and A Saurous, Rif and Kim, Yoon},
  journal = NeurIPS,
  volume  = {36},
  year    = {2024}
}

@misc{espasa2023challengesmodellingsolvingplotting,
  title         = {Challenges in Modelling and Solving Plotting with PDDL},
  author        = {Joan Espasa and Ian Miguel and Peter Nightingale and András Z. Salamon and Mateu Villaret},
  year          = {2023},
  eprint        = {2310.01470},
  archiveprefix = {arXiv},
  primaryclass  = {cs.AI}
}

@article{valmeekam2023planning,
  title   = {On the planning abilities of large language models-a critical investigation},
  author  = {Valmeekam, Karthik and Marquez, Matthew and Sreedharan, Sarath and Kambhampati, Subbarao},
  journal = NeurIPS,
  volume  = {36},
  pages   = {75993--76005},
  year    = {2023}
}

@article{Oswald_Srinivas_Kokel_Lee_Katz_Sohrabi_2024,
  title   = {Large Language Models as Planning Domain Generators},
  volume  = {34},
  doi     = {10.1609/icaps.v34i1.31502},
  journal = ICAPS,
  author  = {Oswald, James and Srinivas, Kavitha and Kokel, Harsha and Lee, Junkyu and Katz, Michael and Sohrabi, Shirin},
  year    = {2024},
  month   = {May},
  pages   = {423-431}
}

@inproceedings{huang2022zerolanguageplan,
  title        = {Language models as zero-shot planners: Extracting actionable knowledge for embodied agents},
  author       = {Huang, Wenlong and Abbeel, Pieter and Pathak, Deepak and Mordatch, Igor},
  booktitle    = ICML,
  pages        = {9118--9147},
  year         = {2022},
  organization = {PMLR}
}

@article{huang2023groundedrobot,
  title   = {Grounded decoding: Guiding text generation with grounded models for robot control},
  author  = {Huang, Wenlong and Xia, Fei and Shah, Dhruv and Driess, Danny and Zeng, Andy and Lu, Yao and Florence, Pete and Mordatch, Igor and Levine, Sergey and Hausman, Karol and others},
  journal = {arXiv preprint arXiv:2303.00855},
  year    = {2023}
}

@misc{zhang2024dkpromptdomainknowledgeprompting,
  title         = {DKPROMPT: Domain Knowledge Prompting Vision-Language Models for Open-World Planning},
  author        = {Xiaohan Zhang and Zainab Altaweel and Yohei Hayamizu and Yan Ding and Saeid Amiri and Hao Yang and Andy Kaminski and Chad Esselink and Shiqi Zhang},
  year          = {2024},
  eprint        = {2406.17659},
  archiveprefix = {arXiv},
  primaryclass  = {cs.AI},
}

@misc{githubollama,
  author = {github},
  title  = {GitHub},
  year   = {2024},
}

@misc{googlegeminifamilyhighlycapable,
  title         = {Gemini: A Family of Highly Capable Multimodal Models},
  author        = {Gemini Team et. al. },
  year          = {2024},
  eprint        = {2312.11805},
  archiveprefix = {arXiv},
  primaryclass  = {cs.CL},
}

@misc{openai2023gpt4vision,
  title         = {OpenAI (2023)},
  author        = {OpenAI},
  year          = {2023},
  eprint        = {2303.08774},
  archiveprefix = {arXiv},
  primaryclass  = {cs.CL},
  url           = {https://openai.com/index/gpt-4v-system-card/}
}

@misc{liu2023llava,
  title     = {Visual Instruction Tuning},
  author    = {Liu, Haotian and Li, Chunyuan and Wu, Qingyang and Lee, Yong Jae},
  publisher = {NeurIPS},
  year      = {2023}
}

@inproceedings{onishchenko2025lookplangraph,
  title     = {LookPlanGraph: Embodied instruction following method with {VLM} graph augmentation},
  author    = {Anatoly Onishchenko and Alexey Kovalev and Aleksandr Panov},
  booktitle = {Workshop on Reasoning and Planning for Large Language Models},
  year      = {2025},
}

@misc{grigorev2025verifyllmllmbasedpreexecutiontask,
  title         = {VerifyLLM: LLM-Based Pre-Execution Task Plan Verification for Robots},
  author        = {Danil S. Grigorev and Alexey K. Kovalev and Aleksandr I. Panov},
  year          = {2025},
  eprint        = {2507.05118},
  archiveprefix = {arXiv},
  primaryclass  = {cs.RO},
}

@misc{pchelintsev2025lerareplanningvisualfeedback,
  title         = {LERa: Replanning with Visual Feedback in Instruction Following},
  author        = {Svyatoslav Pchelintsev and Maxim Patratskiy and Anatoly Onishchenko and Alexandr Korchemnyi and Aleksandr Medvedev and Uliana Vinogradova and Ilya Galuzinsky and Aleksey Postnikov and Alexey K. Kovalev and Aleksandr I. Panov},
  year          = {2025},
  eprint        = {2507.05135},
  archiveprefix = {arXiv},
  primaryclass  = {cs.RO},
}

@software{openclip_ilharco_gabriel_2021,
  author    = {Ilharco, Gabriel and
               Wortsman, Mitchell and
               Wightman, Ross and
               Gordon, Cade and
               Carlini, Nicholas and
               Taori, Rohan and
               Dave, Achal and
               Shankar, Vaishaal and
               Namkoong, Hongseok and
               Miller, John and
               Hajishirzi, Hannaneh and
               Farhadi, Ali and
               Schmidt, Ludwig},
  title     = {OpenCLIP},
  month     = jul,
  year      = 2021,
  note      = {If you use this software, please cite it as below.},
  publisher = {Zenodo},
  version   = {0.1},
  doi       = {10.5281/zenodo.5143773},
}

@article{fang2023anygrasp,
  title   = {AnyGrasp: Robust and Efficient Grasp Perception in Spatial and Temporal Domains},
  author  = {Fang, Hao-Shu and Wang, Chenxi and Fang, Hongjie and Gou, Minghao and Liu, Jirong and Yan, Hengxu and Liu, Wenhai and Xie, Yichen and Lu, Cewu},
  journal = TRO,
  year    = {2023}
}

@misc{medeiros2023langsam,
  author       = {Luca Medeiros},
  title        = {Lang Segment Anything},
  year         = {2023},
  howpublished = {\url{https://github.com/luca-medeiros/lang-segment-anything}}
}

\end{document}